 \documentclass{article}
 \PassOptionsToPackage{numbers, compress}{natbib}
 \usepackage[preprint]{neurips_2025}
 \usepackage{graphicx}
 \usepackage[utf8]{inputenc} % allow utf-8 input
 \usepackage[T1]{fontenc}    % use 8-bit T1 fonts
 \usepackage{hyperref}       % hyperlinks
 \usepackage{url}            % simple URL typesetting
 \usepackage{booktabs}       % professional-quality tables
 \usepackage{amsfonts}       % blackboard math symbols
 \usepackage{nicefrac}       % compact symbols for 1/2, etc.
 \usepackage{microtype}      % microtypography
 \usepackage{xcolor}         % colors
 \usepackage{changepage}  % 在导言区引入
 \usepackage{graphicx}    % 处理图像
 \usepackage{subcaption}  % 更现代的子图环境（推荐）
 \usepackage{caption}     % 控制标题格式
 \usepackage{amsmath} 
 \usepackage{multirow} 
 \usepackage{enumitem}
 \usepackage{tabularx}

 \usepackage[utf8]{inputenc}
 \usepackage{amsmath}
 \usepackage{amsfonts}
 \usepackage{algorithm}
 \usepackage{algpseudocode}
 \usepackage{comment} % 用于添加注释
 \title{Mutual Learning for Hashing: Unlocking Strong Hash Functions from Weak Supervision}

\author{
  \textbf{Xiaoxu Ma}\textsuperscript{1,2} \quad
  \textbf{Runhao Li}\textsuperscript{3} \quad
  \textbf{Zhenyu Weng}\textsuperscript{1}\thanks{Corresponding author.} \\[0.3em] 
  \textsuperscript{1}Shien-Ming Wu School of Intelligent Engineering, South China University of Technology \\
  \textsuperscript{2}School of Electrical and Computer Engineering, Georgia Institute of Technology \\
  \textsuperscript{3}School of Electrical and Electronic Engineering, Nanyang Technological University \\[0.3em]
  \texttt{xma394@gatech.edu}, \texttt{runhao001@e.ntu.edu.sg}, \texttt{wzytumbler@gmail.com}
}

\begin{document}
 \maketitle

  \begin{abstract}
Deep hashing has been widely adopted for large-scale image retrieval, with numerous strategies proposed to optimize hash function learning. Pairwise-based methods are effective in learning hash functions that preserve local similarity relationships, whereas center-based methods typically achieve superior performance by more effectively capturing global data distributions. However, the strength of center-based methods in modeling global structures often comes at the expense of underutilizing important local similarity information. To address this limitation, we propose Mutual Learning for Hashing (MLH), a novel weak-to-strong framework that enhances a center-based hashing branch by transferring knowledge from a weaker pairwise-based branch. MLH consists of two branches: a strong center-based branch and a weaker pairwise-based branch. Through an iterative mutual learning process, the center-based branch leverages local similarity cues learned by the pairwise-based branch. Furthermore, inspired by the mixture-of-experts paradigm, we introduce a novel mixture-of-hash-experts module that enables effective cross-branch interaction, further enhancing the performance of both branches. Extensive experiments demonstrate that MLH consistently outperforms state-of-the-art hashing methods across multiple benchmark datasets.
\end{abstract}

  \section{Introduction}

Efficient image representation is fundamental for large-scale multimedia retrieval \citep{kong2024mitigating,wu2023forb,shao2023global,liang2021deep,shen2021re}. Hashing has emerged as a prominent solution due to its advantages in computation and storage efficiency \citep{li2011hashing,jiang2025online,cao2025deep,liang2024self}, converting high-dimensional visual features into compact binary codes while preserving semantic similarity in the Hamming space \citep{cao2018deep,yang2018adversarial,qin2018gph,jegou2008hamming}. Recent advances in deep learning-based hashing methods have achieved state-of-the-art performance by jointly optimizing feature extraction and hash code generation in an end-to-end fashion \citep{yuan2020central,hoe2021one,wang2023deep,he2024flexible,liu2016deep,li2015feature,cao2017hashnet,wang2017deep,su2018greedy,fan2020deep}.  

Based on the supervision paradigm, deep supervised hashing methods can be broadly categorized into pairwise-based \citep{liu2016deep,li2015feature,cao2017hashnet}, tripletwise-based \citep{wang2017deep,deng2018triplet}, listwise-based \citep{liang2021deep} and center-based \citep{su2018greedy,fan2020deep,yuan2020central,hoe2021one,wang2023deep} approaches. Pairwise methods focus on learning from binary relationships between sample pairs: similar pairs are encouraged to have close hash codes, while dissimilar pairs are pushed apart. Tripletwise methods extend this by modeling relative similarity, optimizing over triplets composed of an anchor, a positive, and a negative sample. Listwise methods consider the ranking order of multiple items and directly optimize global retrieval metrics, making them particularly suitable for preserving complex semantic structures and inter-class relationships at scale.

More recently, center-based methods \citep{yuan2020central,hoe2021one,wang2023deep} introduce learnable hash centers to directly model the global distribution of hash codes, shifting from relative similarity modeling to absolute class-level representation by encouraging intra-class compactness via center-driven objectives. These methods establish representative hash centers for each category and train the network to align the hash codes of individual samples with their corresponding centers. This center-driven design excels at capturing intra-class similarity \citep{wei2022weakly,xuan2021intra,kobayashi2021t,wang2022improving} by enforcing compactness within each class, and it inherently reflects the global data distribution more effectively. As a result, center-based methods often achieve superior performance in large-scale image retrieval tasks, where discriminative power and global structure preservation are crucial. 

Although each paradigm has achieved success, most existing methods rely on a single type of supervision, missing the opportunity to leverage the strengths of complementary approaches. To address this gap, we propose Mutual Learning for Hashing (MLH), a novel weak-to-strong framework that enhances a strong center-based hashing branch by transferring knowledge from a weaker pairwise-based branch via deep mutual learning. MLH consists of two collaborative branches: a center-based branch that learns hash functions guided by predefined hash centers, and a pairwise-based branch that captures local similarities from sample pairs. Through iterative mutual learning, the pairwise branch benefits from the global semantic structure encoded by the center-based branch, while the center-based branch incorporates local similarity cues from the pairwise counterpart. Interestingly, this process not only helps the weaker pairwise branch produce more effective hash functions, but also enables the stronger center-based branch to improve via weak supervision from its peer. 

To facilitate effective inter-branch communication and better align the architecture with hashing objectives, we further propose the Mixture-of-Hash-Experts (MoH) — a customized variant of the Mixture-of-Experts (MoE) \citep{shazeer2017outrageously,chen2023adamv,chen2022towards,riquelme2021scaling} framework, specifically tailored for hash code learning. MoH projects input features into continuous hash codes, treating each expert as a specialized hash layer. To balance consistency and diversity, we employ shared experts that preserve transformation consistency across branches, while allowing each branch to maintain an independent gating mechanism. Together, these components form a unified and principled framework that fully exploits complementary supervision signals for enhanced deep hashing performance.

%To further promote effective cross-branch interaction, we introduce a novel hash layer in both branches inspired by the Mixture-of-Experts (MoE) architecture, which contributes to enhanced performance across the framework. Extensive experiments on standard benchmarks including CIFAR-10, MSCOCO, and ImageNet demonstrate that MLH significantly outperforms state-of-the-art deep hashing methods in terms of retrieval precision.

\begin{figure}[t]
    \centering
    %\hspace*{-0.1\textwidth} % 向左移动 0.1\textwidth，使左右溢出相等
    \includegraphics[width=1.0\textwidth]{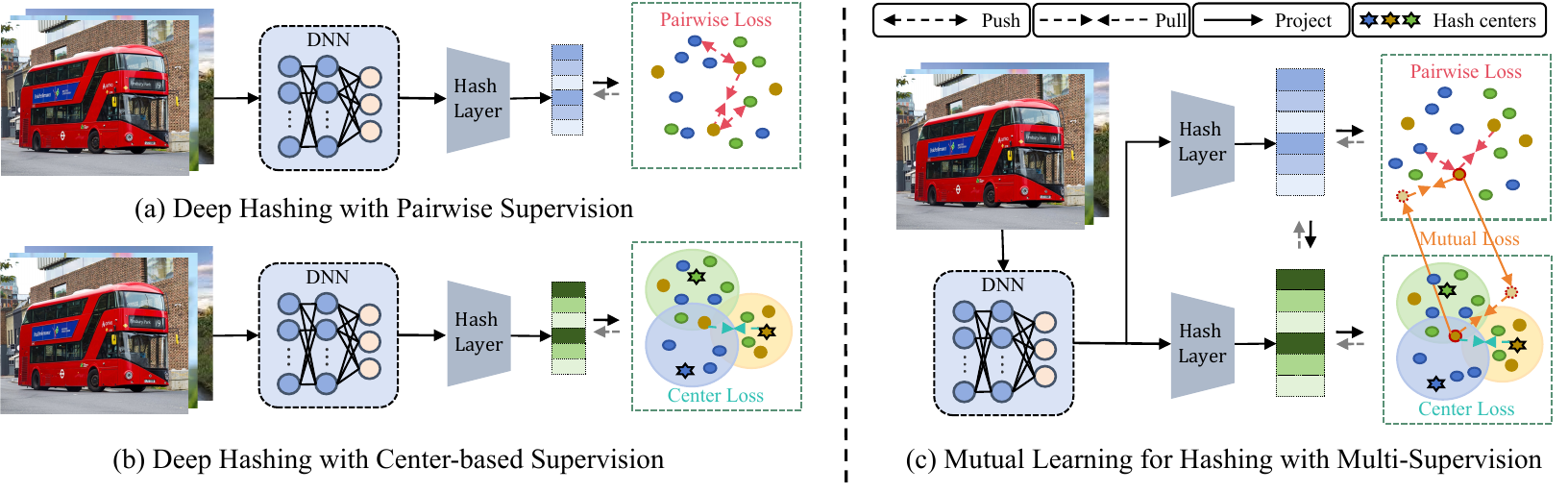}
    \caption{Comparison of hash code learning strategies. (a) Pairwise supervision captures local similarity but lacks global structure. (b) Center-based supervision emphasizes global semantics while ignoring local relations. (c) The proposed MLH integrates both via dual hash layers and deep mutual learning. Cyan, orange, and red arrows indicate center-based, mutual, and pairwise loss, respectively.}
    \label{fig:structure_comparison}
\end{figure}

In summary, our main contributions are summarized as follows:
\begin{itemize}[label=$\bullet$]

    \item We introduce a novel weak-to-strong framework in which a weaker pairwise-based branch guides and enhances the performance of a stronger center-based branch.
    
    \item We present a mutual learning strategy that leverages two distinct supervised paradigms for hashing, enabling them to complement and improve each other.
    
    \item We propose Mixture-of-Hash-Experts (MoH), a hashing-specific module that maps features to the hash space and enables cross-branch communication.
    
    \item Extensive experiments across multiple datasets show that MLH surpasses state-of-the-art deep hashing methods in retrieval precision.

\end{itemize}
  
\section{Related Works}

\noindent\textbf{Weak-to-strong learning} \citep{burns2023weak,lang2024theoretical,zheng2020deep,yang2024weak,gambashidze2024weak} has emerged as a powerful strategy to improve model performance by utilizing weaker models to enhance stronger ones. Burns et al. \citep{burns2023weak} pioneered weak-to-strong generalization in natural language processing, demonstrating that a weaker model can supervise a stronger model through knowledge distillation, using an augmented confidence loss to achieve significant performance gains. In computer vision, Gambashidze et al. \citep{gambashidze2024weak} proposed X-Ray distillation for 3D object detection, where a weaker X-Ray Teacher, trained on object-complete frames, supervises a stronger student model via distillation to address LiDAR point cloud challenges. Our work applies weak-to-strong learning to deep hashing, where a weaker pairwise-based hashing branch can improve a stronger center-based hashing branch through mutual learning. Unlike previous methods, which rely on distillation for supervision, our method uses mutual learning to enable bidirectional learning, allowing the weak hashing branch to enhance the strong hashing branch. 

\noindent\textbf{Deep Mutual Learning (DML)} \citep{zhang2018deep,guo2022online,zhao2021novel,zhao2021deep}, first proposed by Zhang et al. \citep{zhang2018deep}, is a collaborative training strategy where multiple neural networks learn simultaneously by aligning their predicted class probabilities to enhance generalization performance. Extending this idea, Zhao et al. \citep{zhao2021deep} introduced a novel application of deep mutual learning to visual object tracking, leveraging mutual supervision between lightweight networks during offline training to improve backbone representations and tracking accuracy. Unlike previous methods that train multiple networks concurrently using the same type of objective functions, we propose a novel mutual learning framework for deep hashing, in which two branches—each adopting a distinct supervised paradigm—iteratively learn from each other through hash codes similarity to optimize hash functions \citep{cao2018deep,wang2022supervised}.

\noindent\textbf{Mixture-of-Experts (MoE)} \citep{shazeer2017outrageously,chen2023adamv,chen2022towards,riquelme2021scaling}, proposed for large-scale neural networks by Shazeer et al. \citep{shazeer2017outrageously}, employs multiple specialized experts with a top-k gating mechanism for efficient task execution. Chen et al. \citep{chen2023adamv} take the lead in applying Mixture-of-Experts (MoE) to multi-task learning in computer vision by proposing AdaMV-MoE, which introduces task-specific routing and adaptive expert selection to enhance performance across diverse recognition tasks. Inspired by the success of MoE in handling task-specific information, we propose Mixture-of-Hash-Experts (MoH) within our framework. MoH adapts the MoE paradigm to hashing by designing experts that directly project input features into continuous hash codes. We also employ shared experts that preserve transformation consistency across branches, while allowing each branch to maintain an independent gating mechanism

  \section{Methodology}

\begin{figure}[t]
    \centering
    %\hspace*{-0.1\textwidth} % 向左移动 0.1\textwidth，使左右溢出相等
    \includegraphics[width=1.0\textwidth]{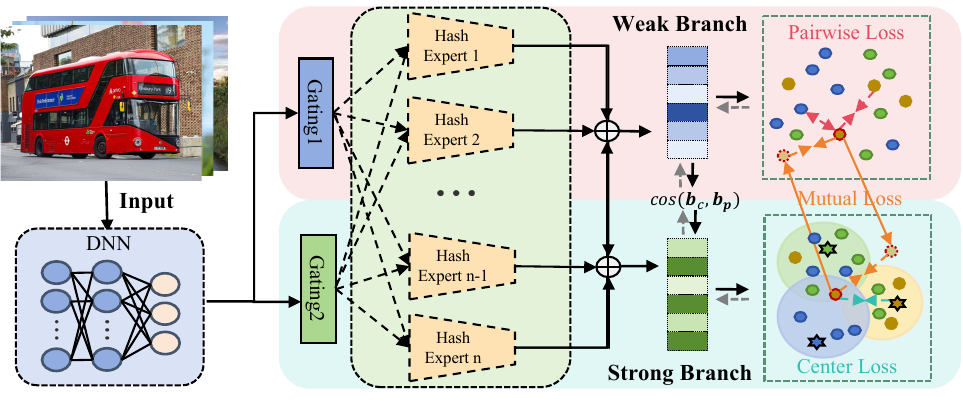}
    \caption{Overview of the proposed Mutual Learning for Hashing (MLH) framework. A deep neural network extracts image features, which are then passed through two parallel branches: a pairwise-supervised weak branch and a center-supervised strong branch. Each branch generates hash codes via its own hash layer, enabling mutual learning between local and global similarity structures.}
    \label{fig:overall_structure}
\end{figure}

We propose MLH, a dual-branch weak-to-strong framework that combines local pairwise similarity and global structure awareness. The two branches share a mixture-of-hash-experts \citep{shazeer2017outrageously,chen2022towards,riquelme2021scaling} module and are optimized separately using pairwise-based \citep{li2015feature} and center-based loss \citep{wang2018cosface, hoe2021one, wang2023deep} functions. Both branches exchange semantic cues through mutual guidance \citep{zhang2018deep}. MLH is jointly optimized with a hybrid loss that balances all components. The following sections describe each module and loss function in detail, with the overall structure illustrated in Figure~\ref{fig:overall_structure}.

\subsection{Problem Formulation}
Given a dataset of images \( X = \{x_1, x_2, \dots, x_N\} \) and their corresponding labels \( Y = \{y_1, y_2, \dots, y_N\} \), image hash learning aims to learn a mapping \( M: X \to \{-1, 1\}^q \) that encodes an image \( x_i \in X \) into a \( q \)-bit binary code, such that semantically similar images yield codes close in Hamming space, while dissimilar ones are mapped farther apart. In deep hashing, \( M \) is typically realized by using a single branch deep neural network that extracts features, which are subsequently passed through a hash layer to generate continuous codes \( \mathbf{u}_i \in (-1, 1)^q \), followed by binarization via the sign function to produce discrete codes \( \mathbf{b}_i \in \{-1, 1\}^q \).

% a deep neural network backbone \( \phi(\cdot) \) first extracts low-level semantic features:
% \[
% V_0 = \phi(X) = \{v_1, v_2, \dots, v_N\}, \quad \text{where } v_n = \phi(x_n).
% \]
  \subsection{Branch-specific Supervision Objectives}

To effectively optimize the hash function $M$ for both global semantic consistency and local similarity preservation, we design MLH as a dual-branch architecture. Each branch learns image hashing from different perspectives: one focusing on class-level alignment, the other on pairwise relationships. This design ensures that the learned hash codes simultaneously reflect global class structure and fine-grained similarity cues.
Each branch of the MLH framework is supervised by a distinct type of loss. Specifically, the weak branch leverages a pairwise-based loss to capture local similarities while the strong branch employs a center-based loss to encourage samples within the same class to be clustered around a shared semantic center.

\noindent\textbf{The center-based branch} uses pre-generated hash centers to enforce global semantic structure by assigning data points from the same class to the same center. In center generation stage, we define $c$ semantic classes, each associated with a unique $q$-bit binary code $\mathbf{h}_i \in \{-1, 1\}^q$ for $1 \leq i \leq c$. To ensure sufficient separation between hash centers, the minimum Hamming distance $d$ is selected based on the Gilbert-Varshamov (GV) bound~\cite{wang2023deep, richelson2023gilbert, varshamov1957estimate}, satisfying:
\begin{equation}
\sum_{i=0}^{d-2} \binom{q}{i} < 2^c \leq \sum_{i=0}^{d-1} \binom{q}{i}.
\end{equation}

In the training stage, we optimize the similarity between an image's continuous hash code \(\mathbf{u}_j\) and its corresponding center. The probability that a sample belongs to class \(i\) is computed via a softmax over scaled cosine similarity:

\begin{equation}
P_{j,i} = \frac{\exp[\sqrt{q} \cos(\mathbf{u}_j^{\text{c}}, \mathbf{h}_i)]}{\sum_{m=1}^{c} \exp[\sqrt{q} \cos(\mathbf{u}_j^{\text{c}}, \mathbf{h}_m)]},
\end{equation}

where \(q\) is  the length of hash codes and \(\cos(\mathbf{x}, \mathbf{y})\) denotes the cosine similarity. The center-based loss is a cross-entropy formulation:

\begin{equation}
L_C = -\frac{1}{N} \sum_{j=1}^{N} \sum_{i=1}^{c} y_{j,i} \log P_{j,i} + (1 - y_{j,i}) \log (1 - P_{j,i}).
\end{equation}

This encourages alignment of each sample with its target center and separation from others, enhancing intra-class discriminability.

\noindent\textbf{The pairwise-based branch} captures local structure by modeling semantic similarity between pairs. A similarity matrix \(S_{ij} = \mathbb{I}(\mathbf{y}_i^\top \mathbf{y}_j > 0)\) is constructed, and similarity scores are computed via the inner product of hash vectors:

\begin{equation}
I_{ij} = \frac{1}{2} \left( \mathbf{u}_i^{\text{p}} \right)^\top \mathbf{u}_j^{\text{p}}
\end{equation}

The pairwise loss encourages similarity for positive pairs and separation for negative ones:

\begin{equation}
L_P = \frac{1}{N} \sum_{i,j} \left[ \log (1 + e^{-|I_{ij}|}) + \max(0, I_{ij}) - S_{ij} I_{ij} \right].
\end{equation}

Together, these two branches enable the network to learn hash codes that balance global compactness with fine-grained local similarity, improving retrieval performance across diverse tasks.
  \subsection{Cross-Branch Deep Mutual Learning}

Inspired by DML~\citep{zhang2018deep, wu2019mutual}, we aim to enable cross-branch knowledge transfer, allowing the pairwise-based branch to guide the center-based branch through mutual supervision. This strategy enhances global semantic representations with fine-grained local cues.

Let \(\mathbf{u}^{\text{c}}\) and \(\mathbf{u}^{\text{p}}\) denote the continuous hash codes produced by the center-based and pairwise branches, respectively. The mutual learning objective is formulated as a cosine-based mutual loss:
\begin{equation}
L_\text{M} = \mathbb{E}_{(x)}\left[1 - \cos\left(\mathbf{u}^{\text{c}}, \mathbf{u}^{\text{p}}\right)\right].
\end{equation}

% Describing the alternating distillation strategy
To ensure stable and effective knowledge exchange, we alternate the learning direction across training epochs by swapping the detached and optimized term. Formally, this can be expressed as:
\begin{equation}
L_\text{M} = \begin{cases} 
\mathbb{E}\left[1 - \cos\left(\mathbf{u}^{\text{p}}, \text{stop\_grad}(\mathbf{u}^{\text{c}})\right)\right], & \text{if } \text{epoch} \bmod 2 = 0, \\ 
\mathbb{E}\left[1 - \cos\left(\mathbf{u}^{\text{c}}, \text{stop\_grad}(\mathbf{u}^{\text{p}})\right)\right], & \text{if } \text{epoch} \bmod 2 = 1. 
\end{cases}
\end{equation}

% Explaining the benefits of the alternating scheme
This alternating scheme allows the center-based branch to consistently benefit from the auxiliary pairwise-based branch, preserving global discrimination while integrating local consistency.

  \subsection{Overall Objective Function}

The overall objective combines three loss functions:
\begin{equation}
L = \lambda_1 L_C + \lambda_2 L_P + \lambda_3 L_M \label{eq:loss}
\end{equation}
where $\lambda_1$, $\lambda_2$ and $\lambda_3$ are trade-off hyperparameters. We found the optimal combination to be $\lambda_1 = 4$, $\lambda_2 = 1$, and $\lambda_3 = 1$ and the detailed tuning experiments are provided in the appendix.

\subsection{ Mixture-of-Hash-Experts (MoH): A Hashing-Oriented Expert Module}

To further enhance the cross-branch interaction, we introduce a novel Mixture of Hash Experts (MoH) module into both branches. MoH is a task-specific variant of the classic Mixture of Experts (MoE) architecture, tailored by us for deep hashing. 

\begin{figure}[t]
    \centering
    %\hspace*{-0.1\textwidth} % 向左移动 0.1\textwidth，使左右溢出相等
    \includegraphics[width=1.0\textwidth]{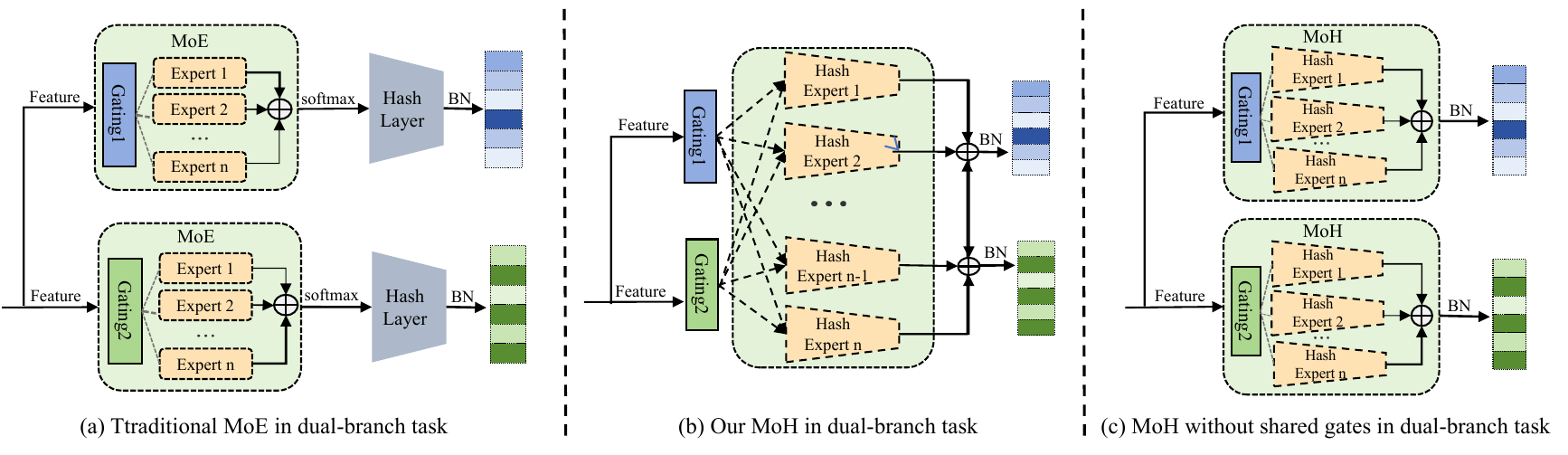}
    \caption{Comparison of expert-based hashing architectures. (a) Traditional mixture-of-experts (MoE) used in dual-branch tasks with separate expert modules. (b) The proposed Mixture of Hashing Experts (MoH), featuring shared experts and independent gates, where each expert generates continuous hash codes. (c) A MoH variant without expert sharing, maintaining the expert-to-hash mapping.}
    \label{fig:MoH}
\end{figure}

Given a set of input images \( X \), a deep neural network backbone \( \phi(\cdot) \) is first used to extract shared semantic features:
\begin{equation}
V_0 = \phi(X) = \{\mathbf{v}_1, \mathbf{v}_2, \dots, \mathbf{v}_N\}, \quad \text{where } \mathbf{v}_n = \phi(x_n).
\end{equation}

To better accommodate the unique learning objectives of each branch, we design two separate gating networks \( G^{\text{c}} \) and \( G^{\text{p}} \), corresponding to the center-based and pairwise-based streams respectively. These gating networks dynamically determine expert activation for each input based on the shared features \( v_n \). A set of shared experts \( \{E_i\}_{i=1}^m \), each implemented as a lightweight neural network, project the input feature directly into the continuous hash code space \( \mathbb{R}^q \):
\begin{equation}
E_i: \mathbb{R}^d \rightarrow \mathbb{R}^q.
\end{equation}

The branch-specific refined representation is then computed as:
\begin{equation}
\mathbf{u}_n^s = \sum_{i=1}^m G^s(\mathbf{v}_n)_i \cdot E_i(\mathbf{v}_n), \quad s \in \{\text{c}, \text{p}\}, \quad \mathbf{u}_n^s \in \mathbb{R}^K.
\end{equation}

Unlike conventional MoE designs that separate feature transformation and task-specific heads shown in Figure~\ref{fig:MoH}(a), MoH treats each expert as a direct generator of semantically meaningful hash codes. This integration shown in Figure~\ref{fig:MoH}(b) and (c) simplifies the architecture and allows each expert to act as a specialized hashing pathway. In Sec. 4.2.2, we will provide an analysis of MoH design. The final binary hash codes are then obtained via:
\begin{equation}
\mathbf{b}_n^s = \operatorname{sign}(\mathbf{u}_n^s).
\end{equation}

By assigning separate gating functions to each branch while sharing the expert pool, our method in Figure~\ref{fig:MoH}(b) encourages diversified yet coordinated learning, enabling the branches to exploit complementary semantics without enforcing rigid alignment. This setup acts as an implicit communication mechanism, bridging inter-branch semantic gaps and enhancing the discriminability of the resulting hash codes. The pseudo-code for MLH is presented in Algorithm 1.

\begin{algorithm}
\caption{MLH: Mutual Learning with MoH}
\begin{algorithmic}[1]
\Require Training set $\mathcal{D}$, number of experts $m$, hash length $K$, iterations $T$, weights $\lambda_1, \lambda_2, \lambda_3$
\State \textbf{Initialize:} DNN backbone $\phi(\cdot)$, shared experts $\{E_i\}_{i=1}^m$, gating networks $G^{\text{c}}, G^{\text{p}}$, hash centers $\{\mathbf{h}_i\}_{i=1}^c$
\For{$t = 1$ to $T$}
    \State Sample a mini-batch $X = \{x_1, \dots, x_N\}$ from $\mathcal{D}$
    \State Extract base features $V_0 = \{\mathbf{v}_1, \dots, \mathbf{v}_N\} \gets \phi(X)$

    \For{each stream $s \in \{\text{c}, \text{p}\}$}
        \For{each sample $v_n$ in $V_0$}
            \State Compute expert outputs: $\{E_i(\mathbf{v}_n)\}_{i=1}^m$
            \State Get gating weights: $\alpha_n^s = G^s(\mathbf{v}_n)$
            \State Refined continuous hash codes: $\mathbf{u}_n^s = \sum_{i=1}^m \alpha_n^s[i] \cdot E_i(\mathbf{v}_n)$
        \EndFor
    \EndFor

    \State Compute center loss $L_C$ using $\{\mathbf{u}_n^{\text{c}}\}$ and label centers $\{\mathbf{h}_i\}$
    \State Compute pairwise loss $L_P$ from $\{\mathbf{u}_n^{\text{p}}\}$ and pairwise similarities

    \State \textbf{Compute mutual loss:}
    \If{$t \bmod 2 = 0$}
        \State Detach $u_n^{\text{p}}$ as target
        \State $L_M = \frac{1}{N} \sum_{n=1}^N \left[1 - \cos(\mathbf{u}_n^{\text{c}}, \text{detach}(\mathbf{u}_n^{\text{p}}))\right]$
    \Else
        \State Detach $u_n^{\text{c}}$ as target
        \State $L_M = \frac{1}{N} \sum_{n=1}^N \left[1 - \cos(\text{detach}(\mathbf{u}_n^{\text{c}}), \mathbf{u}_n^{\text{p}})\right]$
    \EndIf

    \State Total loss: $L \gets \lambda_1 L_C + \lambda_2 L_P + \lambda_3 L_M$
    \State Update $\phi$, $\{E_i\}$, $G^{\text{c}}, G^{\text{p}}$ via RMSProp using $L$
\EndFor
\State \Return Trained model: $\phi$, $\{E_i\}$, $G^{\text{c}}, G^{\text{p}}$
\end{algorithmic}
\end{algorithm}

  \section{Experiments}
\textbf{Datasets.} Following prior works \citep{yuan2020central,hoe2021one,wang2023deep,liu2016deep,li2015feature,cao2017hashnet,wang2017deep,su2018greedy,fan2020deep}, we evaluate performance on CIFAR10 \citep{krizhevsky2009learning}, ImageNet \citep{deng2009imagenet}, and MSCOCO \citep{lin2014microsoft} for category-level retrieval. Evaluation metrics include mean average precision (mAP) and precision-recall curves. We report mAP@1000 for CIFAR10 and ImageNet, and mAP@5000 for MSCOCO.

\textbf{Training Setup.} Following \citep{yuan2020central,hoe2021one,wang2023deep}, we use a pre-trained ResNet-50 \citep{he2016deep} as the backbone network $\phi(\cdot)$, extracting 4096-dimensional base features from the final fully-connected ReLU layer \citep{agarap2019deep}. These features are processed by a Mixture-of-Hashing (MoH) module, consisting of $m$ shared expert networks $\{E_i\}_{i=1}^m$, two gating networks $G^{\text{c}}$ and $G^{\text{p}}$, and hash centers $\{\mathbf{h}_i\}_{i=1}^c$ for the center stream. Input images are resized to 224$\times$224, and we use a mini-batch size of $N=64$. The model is trained for $T=100$ iterations (epochs) to optimize the backbone $\phi$, experts $\{E_i\}$, and gating networks $G^{\text{c}}$, $G^{\text{p}}$, by jointly optimizing $\lambda_1$, $\lambda_2$, and $\lambda_3$. The final binary hash code is obtained as $\text{sign}(\mathbf{u}_n^s)$, where $\mathbf{u}_n^s$ ($s \in \{\text{c}, \text{p}\}$) are the refined continuous hash codes from the center and pairwise streams. Our model is implemented in PyTorch and trained on an NVIDIA RTX 4090 GPU using the RMSProp optimizer with a learning rate of 0.0001.

  \subsection{Results of Retrieval Accuracy}

\begin{table}[t]
    \centering
    \caption{Comparison results of retrieval performance w.r.t. mAP on three datasets across different bit configuration. 
    \textbf{Bold} values indicate the best performance, and \underline{underlined} values indicate the second best performance.}
    \renewcommand{\arraystretch}{1.2}
    \resizebox{\textwidth}{!}{ % 缩放到页面宽度
        \begin{tabular}{l ccc ccc ccc}
            \toprule
            \multirow{2}{*}{\textbf{Method}} & \multicolumn{3}{c}{\textbf{CIFAR-10(@1000)}} & \multicolumn{3}{c}{\textbf{ImageNet(@1000)}} & \multicolumn{3}{c}{\textbf{MSCOCO(@5000)}} \\
            \cmidrule(lr){2-4} \cmidrule(lr){5-7} \cmidrule(lr){8-10}
            & 16 bits & 32 bits & 64 bits & 16 bits & 32 bits & 64 bits & 16 bits & 32 bits & 64 bits \\
            \midrule
            DSH\citep{liu2016deep}  & 0.7313 & 0.7402 & 0.7272 & 0.7179 & 0.7448 & 0.7585 & 0.7221 & 0.7573 & 0.7790 \\
            DPSH\citep{li2015feature}  & 0.3098 & 0.3632 & 0.3638 & 0.6241 & 0.7626 & 0.7992 & 0.6239 & 0.6467 & 0.6322 \\
            HashNet\citep{cao2017hashnet}  & 0.8959 & 0.9115 & 0.8995 & 0.6024 & 0.7158 & 0.8071 & 0.7540 & 0.7331 & 0.7882 \\
            DTSH\citep{wang2017deep}  & 0.7783 & 0.7997 & 0.8312 & 0.6606 & 0.7803 & 0.8120 & \underline{0.7702} & 0.8105 & 0.8233 \\
            GreedyHash\citep{su2018greedy}  & 0.3519 & 0.5350 & 0.6177 & 0.7394 & 0.7977 & 0.8243 & 0.7625 & 0.8033 & 0.8570 \\
            DPN\citep{fan2020deep}  & 0.7576 & 0.7901 & 0.8040 & 0.7987 & 0.8298 & 0.8394 & 0.7571 & 0.8227 & 0.8623 \\
            CSQ\citep{yuan2020central}  & 0.7861 & 0.7983 & 0.7989 & 0.8377 & 0.8750 & 0.8836 & 0.7509 & \underline{0.8471} & \underline{0.8610} \\
            OrthoHash\citep{hoe2021one}  & 0.9087 & 0.9297 & 0.9454 & 0.8540 & 0.8792 & 0.8936 & 0.7174 & 0.7675 & 0.8060 \\
            MDSH\citep{wang2023deep}  & \underline{0.9455} & \underline{0.9554} & \underline{0.9607} & \underline{0.8639} & \underline{0.8863} & \underline{0.9019} & 0.7542 & 0.8131 & 0.8143 \\
            \midrule
            \textbf{Ours}  & \textbf{0.9665} & \textbf{0.9657} & \textbf{0.9658} & \textbf{0.8744} & \textbf{0.8975} & \textbf{0.9062} & \textbf{0.7903} & \textbf{0.8675} & \textbf{0.8727} \\
            \bottomrule
        \end{tabular}
    }
    \label{tab:sota_compare}
\end{table}

We compare our method with nine representative deep hashing algorithms: five pointwise methods (DPN \citep{fan2020deep}, GreedyHash \citep{su2018greedy}, CSQ \citep{yuan2020central}, OrthoHash \citep{hoe2021one}, MDSH \citep{wang2023deep}), three pairwise methods (DSH \citep{liu2016deep}, DPSH \citep{li2015feature}, HashNet \citep{cao2017hashnet}), and one tripletwise method (DTSH \citep{wang2017deep}). Table~\ref{tab:sota_compare} reports the Mean Average Precision (mAP) results for image retrieval. We adopt ResNet-50 as the backbone for all compared methods, including DSH, DPSH, DTSH, HashNet, GreedyHash, DPN, CSQ, OrthoHash, MDSH, and our proposed MLH. Compared to state-of-the-art deep hashing approaches, our method achieves mAP improvements of 1.74\%, 1.21\%, and 0.87\% on MSCOCO, CIFAR-10, and ImageNet, respectively, averaged across different code lengths. 

\begin{figure}[t]
    \centering
    \begin{subfigure}[t]{0.33\textwidth}
        \centering
        \includegraphics[width=\textwidth]{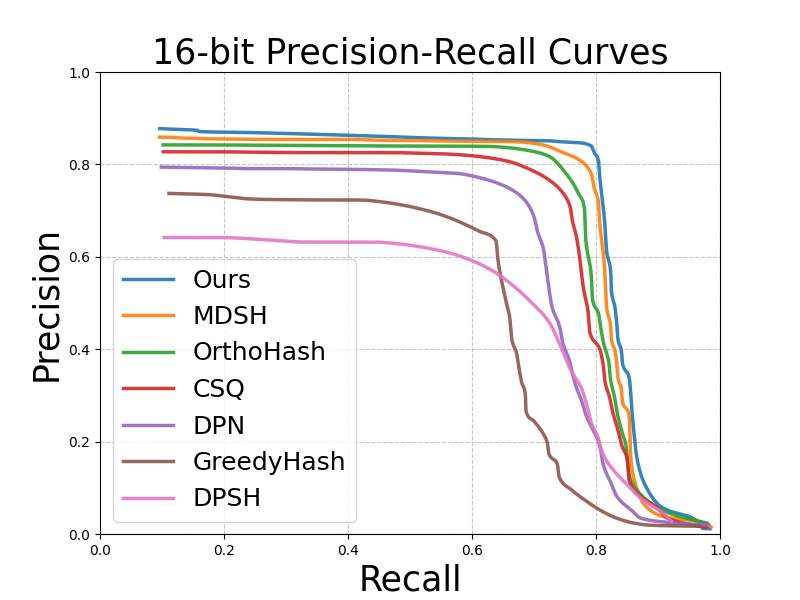}
        \caption{16 bit Precision Recall Curve}
    \end{subfigure}\hfill
    \begin{subfigure}[t]{0.33\textwidth}
        \centering
        \includegraphics[width=\textwidth]{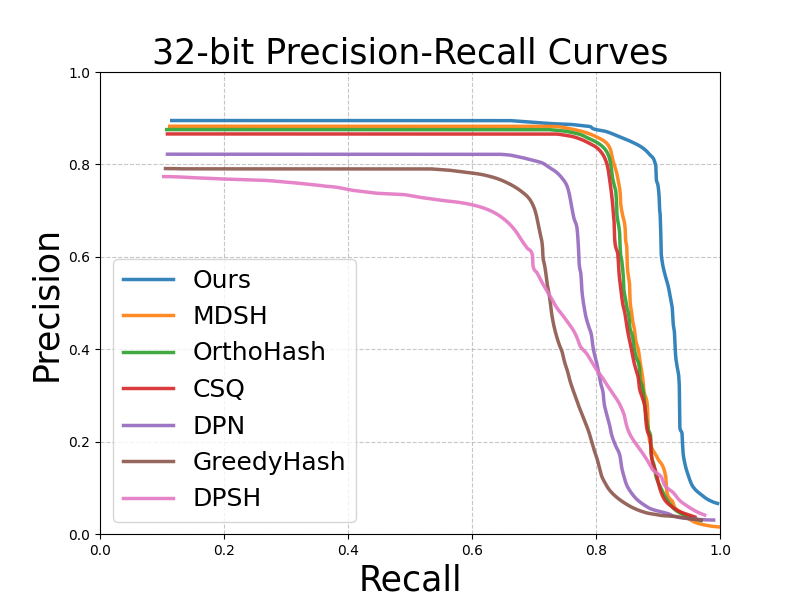}
        \caption{32 bit Precision Recall Curve}
    \end{subfigure}\hfill
    \begin{subfigure}[t]{0.33\textwidth}
        \centering
        \includegraphics[width=\textwidth]{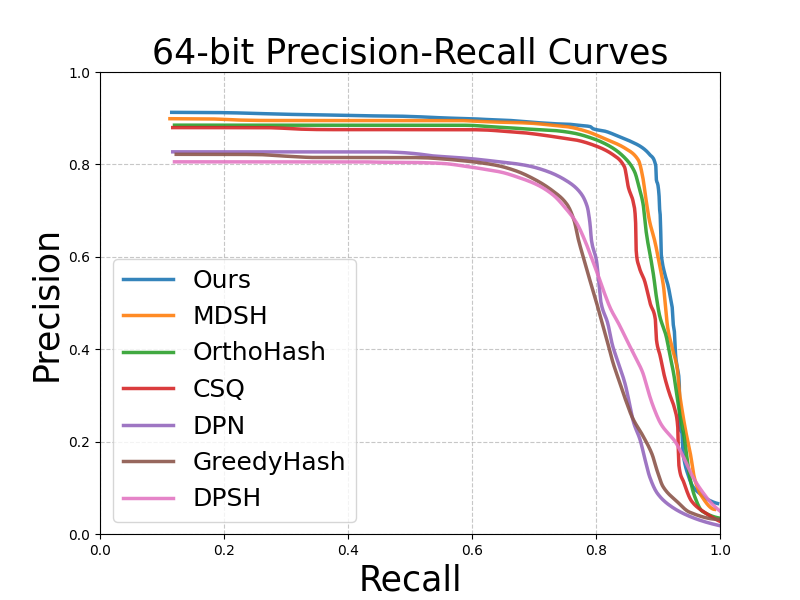}
        \caption{32 bit Precision Recall Curve}
    \end{subfigure}
    \caption{Precision recall curves on ImageNet across different bit configurations.}
    \label{fig:pr_curve}
\end{figure}

We further evaluate retrieval performance using Precision-Recall (PR) curves, as shown in Figure~\ref{fig:pr_curve}. Our method consistently yields a larger Area Under the PR Curve (AUC-PR) across all bit lengths, demonstrating superior precision across a wide range of recall values. These results underscore the robustness and generalization capability of our method for large-scale image retrieval tasks.

  \subsection{Ablation Study}

We conduct an ablation study on three datasets under different hash code lengths (16, 32, and 64 bits) to evaluate the effectiveness of the proposed components in our deep hashing network, including both the overall framework modules in Table~\ref{tab:ablation} and the detailed design choices of the Mixture-of-Hashing-Experts (MoH) head in Table~\ref{tab:model_comparison}.

\subsubsection{Analysis of Overall Framework}

% \begin{table}[t]
% \centering
% \caption{Ablation study of DML and MoE for MSCOCO across all bit configurations. The center-based and pairwise columns under each bit length indicate the performance of the corresponding branches for each module configuration listed on the left. The best results are \textbf{bolded}.}
% \label{tab:ablation}
% \begin{tabular}{l *{6}{c}}
% \toprule
%  & \multicolumn{2}{c}{16 bits} & \multicolumn{2}{c}{32 bits} & \multicolumn{2}{c}{64 bits} \\
% \cmidrule(lr){2-3} \cmidrule(lr){4-5} \cmidrule(lr){6-7}
% Module & center-based & pairwise & center-based & pairwise & center-based & pairwise \\
% \midrule
% None       & 0.7641 & 0.7008 & 0.8308 & 0.7542 & 0.8475 & 0.7139 \\
% DML      & 0.7781 & 0.7473 & 0.8330 & 0.8223 & 0.8513 & 0.8418 \\
% MoE      & 0.7440 & 0.6571 & 0.8345 & 0.6718 & 0.8295 & 0.6421 \\
% MoE+DML  & \textbf{0.7792} & \textbf{0.7719} & \textbf{0.8617} & \textbf{0.8476} & \textbf{0.8784} & \textbf{0.8682} \\
% \bottomrule
% \end{tabular}
% \end{table}

\begin{table}[t]
\centering
\caption{Ablation study of Mutual learning (ML) and MoH across different datasets for 64bits. The center-based and pairwise columns under each dataset indicate the performance of the corresponding branches for each module configuration. The best results are \textbf{bolded}.}
\label{tab:ablation}
%\small % Reduce font size
\setlength{\tabcolsep}{4pt} % Reduce column spacing (default is 6pt)
\begin{tabular}{ccc *{6}{c}}
\toprule
 & \multicolumn{2}{c}{Modules} & \multicolumn{2}{c}{\textbf{CIFAR-10(@1000)}} & \multicolumn{2}{c}{\textbf{ImageNet(@1000)}} & \multicolumn{2}{c}{\textbf{MSCOCO(@5000)}} \\
\cmidrule(lr){2-3} \cmidrule(lr){4-5} \cmidrule(lr){6-7} \cmidrule(lr){8-9}
Baseline & ML & MoH & center-based & pairwise & center-based & pairwise & center-based & pairwise \\
\midrule
$\checkmark$ &  &  & 0.9607 & 0.9605 & 0.8940 & 0.8873 & 0.8475 & 0.8418 \\
$\checkmark$ & $\checkmark$ &  & 0.9586 & 0.9634 & 0.9037 & 0.8862 & 0.8427 & 0.8605 \\
$\checkmark$ &  & $\checkmark$ & 0.8757 & 0.9655 & 0.8325 & 0.9003 & 0.7139 & 0.8708 \\
$\checkmark$ & $\checkmark$ & $\checkmark$ & \textbf{0.9611} & \textbf{0.9655} & \textbf{0.8997} & \textbf{0.9003} & \textbf{0.8727} & \textbf{0.8513} \\
\bottomrule
\end{tabular}
\end{table}

% \begin{table}[t]
% \centering
% \caption{Ablation study of DML and MoE for MSCOCO across all bit configurations. The center-based and pairwise columns under each bit length indicate the performance of the corresponding branches for each module configuration. The best results are \textbf{bolded}.}
% \label{tab:ablation}
% \small % Reduce font size
% \setlength{\tabcolsep}{4pt} % Reduce column spacing (default is 6pt)
% \begin{tabular}{ccc *{6}{c}}
% \toprule
%  & \multicolumn{2}{c}{Modules} & \multicolumn{2}{c}{16 bits} & \multicolumn{2}{c}{32 bits} & \multicolumn{2}{c}{64 bits} \\
% \cmidrule(lr){2-3} \cmidrule(lr){4-5} \cmidrule(lr){6-7} \cmidrule(lr){8-9}
% Baseline & DML & MoE & center-based & pairwise & center-based & pairwise & center-based & pairwise \\
% \midrule
% $\checkmark$ &  &  & 0.8475 & 0.7139 & 0.8605 & 0.8597 & 0.8727 & 0.8708 \\
% $\checkmark$ & $\checkmark$ &  & 0.8427 & 0.6923 & 0.8513 & 0.8418 & 0.8605 & 0.8597 \\
% $\checkmark$ &  & $\checkmark$ & 0.7440 & 0.6571 & 0.8345 & 0.6718 & 0.8295 & 0.6421 \\
% $\checkmark$ & $\checkmark$ & $\checkmark$ & \textbf{0.7792} & \textbf{0.7719} & \textbf{0.8617} & \textbf{0.8476} & \textbf{0.8784} & \textbf{0.8682} \\
% \bottomrule
% \end{tabular}
% \end{table}

\begin{table}[t]
\centering
\caption{Ablation study of dual-branch MoH components across different datasets for 64bits. The best results are \textbf{bolded}.}
\label{tab:model_comparison}
\small
% \normalsize % Use default font size (already the default, so not needed)
\setlength{\tabcolsep}{6pt} % Increase column spacing to make the table wider
\renewcommand{\arraystretch}{1.0} % Increase row spacing to make the table taller
\begin{tabular}{llcccc}
\toprule
Model Structure & Experts & softmax & \textbf{CIFAR-10(@1000)} & \textbf{ImageNet(@1000)} & \textbf{MSCOCO(@5000)} \\
\midrule
2DNNs+MoE & separate & $\checkmark$ & 0.9634 & 0.8862 & 0.8605 \\
1DNN+MoE  & separate & $\checkmark$ & 0.9643 & 0.9003 & 0.8693 \\
1DNN+MoH  & separate & $\times$     & 0.9651 & 0.9033 & 0.8697 \\
1DNN+MoH  & shared   & $\checkmark$ & 0.9647 & 0.9012 & 0.8684 \\
1DNN+MoH  & shared   & $\times$     & \textbf{0.9658} & \textbf{0.9062} & \textbf{0.8727} \\
\bottomrule
\end{tabular}
% If the table still doesn't fill the page, you can uncomment the following line to scale it:
% \resizebox{\textwidth}{!}{\begin{tabular}{llcccc} ... \end{tabular}}
\end{table}

\begin{figure}[t]
\centering
    \begin{subfigure}[b]{0.33\textwidth}
        \centering
        \includegraphics[width=\textwidth]{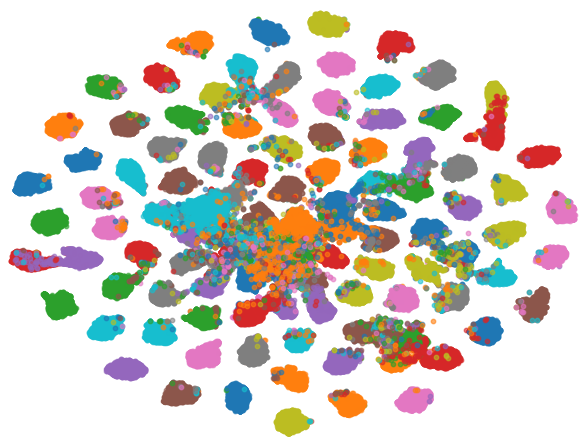}
        \caption{Center-based Method}
    \end{subfigure}\hfill
    \begin{subfigure}[b]{0.33\textwidth}
        \centering
        \includegraphics[width=\textwidth]{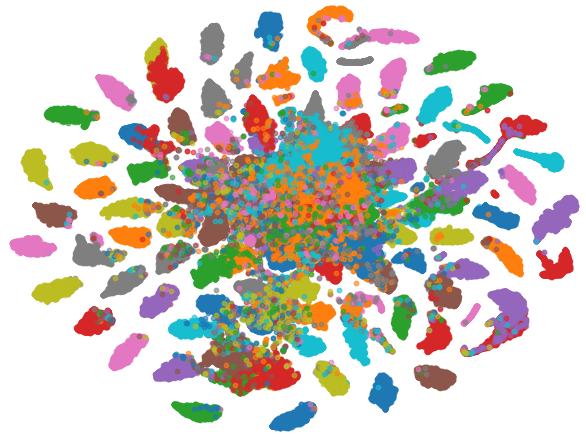}
        \caption{Pairwise-based Method}
    \end{subfigure}\hfill
    \begin{subfigure}[b]{0.33\textwidth}
        \centering
        \includegraphics[width=\textwidth]{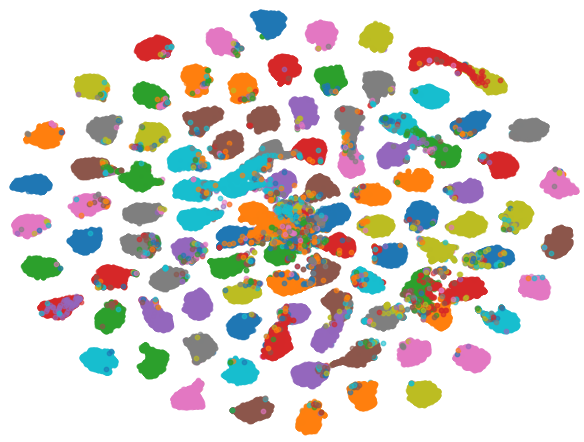}
        \caption{Mutual Learning Method}
    \end{subfigure}
\caption{Impact of MLH on hash code distribution for ImageNet100 with 16-bit configurations. (a) Employs only center-based supervision, focusing on global data structure. (b) Utilizes solely pairwise-based supervision, emphasizing local similarity relationships. (c) Integrates mutual learning supervision, enabling the weak pairwise branch to fine-tune the strong center-based branch for improved hash code optimization.}
\label{fig:comparison}
\end{figure}

% \textbf{The baseline model (Baseline)}, which includes none of MoE,MoH and DML, shows that the center-based branch consistently outperforms the pairwise branch (e.g., 0.8475 vs. 0.7139 at 64 bits), indicating its superior standalone effectiveness.

% \textbf{Introducing DML alone} leads to consistent improvements over the baseline, particularly for the weaker pairwise branch (e.g., 0.8418 vs. 0.7139 at 16 bits), suggesting that DML enhances supervision by enabling bidirectional knowledge transfer. The center-based branch also benefits, though to a lesser extent (e.g., 0.8513 vs. 0.8475), likely due to the still limited diversity between the two branches when MoE is not present.

% \textbf{Incorporating the MoH modules alone} results in a slight performance drop in both branches (e.g., center-based: 0.8427 vs. 0.8475; pairwise: 0.6923 vs. 0.7139 at 16 bits). While MoH introduces structured cross-branch interaction, its effect remains limited without collaborative training, and the exchanged information may not be fully leveraged.

% \textbf{Combining MoH with DML} leads to the best overall performance across all settings (e.g., 0.8727 center-based and 0.8708 pairwise at MSCOCO; 0.9062 and 0.9003 at 64 ImageNet). MoH facilitates effective communication between the center-based and pairwise branches while preserving their individual strengths, and DML further amplifies this interaction through mutual supervision. The synergy between the two yields enhanced intra-class compactness and inter-class separability.

To evaluate the contribution of each component in our overall architecture, we conduct an ablation study on both the MoH and Mutual Learning(ML) modules, as shown in Table~\ref{tab:ablation}.

\textbf{Baseline} (without either MoH or ML) shows that the center-based branch consistently outperforms the pairwise branch (e.g., 0.8475 vs. 0.7139 at 64 bits), indicating stronger standalone effectiveness. \textbf{ML alone} improves both branches, especially the weaker pairwise one (e.g., 0.8418 vs. 0.7139 at 16 bits), by enabling bidirectional knowledge transfer. Gains for the center-based branch are smaller due to limited diversity without MoE. \textbf{MoH alone} slightly reduces performance (e.g., center-based: 0.8427 vs. 0.8475), suggesting that without collaborative training, the structured interaction it introduces is underutilized. \textbf{MoH + ML} achieves the best results (e.g., 0.8727/0.8708 on MSCOCO, 0.9062/0.9003 on ImageNet at 64 bits), as MoH enhances inter-branch communication and mutual learning amplifies mutual supervision, improving both intra-class compactness and inter-class separability.

\subsubsection{Analysis of Mixture-of-Hash-Experts (MoH)}

To further demonstrate the superiority of our proposed MoH module, we compare it with the traditional MoE and perform ablation experiments on two key design choices within MoH: using shared experts and removing the softmax. Additionally, to justify our use of a single DNN instead of the conventional dual-DNN setup in mutual learning, we include a comparison with the two-DNN baseline. The results are presented in Table~\ref{tab:model_comparison}.

\textbf{The traditional mutual learning setup (2DNNs+MoE)} employs two separate DNNs with expert modules. It achieves noticeable improvements over the baseline (e.g., 0.9643 vs. 0.9607 on CIFAR-10), demonstrating the benefit of collaborative learning. However, to further enhance performance and simplify the architecture, we explore alternative designs. \textbf{Switching to a single-branch structure (1DNN+MoE)} leads to slightly improved results (e.g., 0.9634 vs. 0.9643 on CIFAR-10; 0.8605 vs. 0.8693 on MSCOCO), confirming that a unified backbone provides more coherent feature learning for hashing tasks. \textbf{Replacing MoE with our proposed MoH module}—which maps features directly to the hash space—further improves performance (e.g., 0.9651 on CIFAR-10, 0.9033 on ImageNet), demonstrating its better alignment with deep hashing objectives. This gain holds for both separate and shared expert configurations. \textbf{Sharing experts across branches} removes redundancy and enhances learning consistency, with shared MoH slightly outperforming its separate counterpart (e.g., 0.9647 vs. 0.9651 on CIFAR-10; 0.8684 vs. 0.8697 on MSCOCO). \textbf{Removing the softmax layer} yields the best performance across all datasets (e.g., 0.9658 on CIFAR-10, 0.9062 on ImageNet, 0.8727 on MSCOCO), likely due to improved code separability by avoiding over-smoothing among experts.

\subsection{Hash Codes Visualization}

To illustrate the effectiveness of our proposed Mutual Learning for Hashing (MLH) framework, we visualize the t-SNE \citep{donahue2013decaf} of hash codes generated by three methods on ImageNet100, as shown in Figure~\ref{fig:comparison}: (a) center-based \citep{wang2023deep}, (b) pairwise-based \citep{zheng2020deep}, and (c) our MLH approach.

In Figure~\ref{fig:comparison}(a), the center-based method produces a circular distribution, indicating decent global alignment but limited fine-grained separation. The pairwise-based method in (b) yields a more elongated structure with more overlapping clusters, suggesting weaker overall structure. In contrast, the MLH approach in (c) leads to a more compact and well-clustered distribution, with fewer ambiguous points across class boundaries.

These results demonstrate that mutual learning effectively enhances intra-class compactness and inter-class separability by allowing center- and pairwise-based branches to complement and refine each other. Consequently, MLH achieves stronger and more discriminative hash representations.

  \section{Conclusions}

In this work, we introduced Mutual Learning for Hashing (MLH), a novel weak-to-strong framework that unifies center-based and pairwise-based hashing through collaborative mutual learning. By integrating mutual learning with Mixture-of-Hash-Experts heads, MLH effectively captures both global semantics and local similarities, unlocking strong discriminative representations from weak supervision. Experiments on multiple benchmarks show consistent gains in mAP, confirming the benefits of mutual learning and expert diversity. These results highlight the potential of mutual learning in deep hashing to bridge heterogeneous objectives and improve representation quality for scalable, robust retrieval systems.

  \clearpage
  \bibliographystyle{unsrtnat}
  \bibliography{references}

  \clearpage
  \appendix

\section{Summary}

This appendix provides detailed insights and additional experimental results to support the main paper.

\section{Training Setup}

\subsection{Datasets}

\textbf{ImageNet} is a large-scale image classification dataset consisting of over 1.2 million images annotated with 1,000 categories. Following the protocol in~\citep{deng2009imagenet}, we use the ILSVRC2012 version for evaluation. The validation set of 50K images is used as the query set, while the remaining training images form the database. For training, we randomly sample 130K images from the database.

\textbf{CIFAR-10} consists of 60,000 images across 10 categories, with each image sized $32\times32$. Following the standard practice in~\citep{krizhevsky2009learning}, we use 10K test images as the query set and the remaining 50K training images as the database. For training, 5K images are randomly sampled from the database.

\textbf{MSCOCO} ~\citep{lin2014microsoft} is an image recognition, segmentation, and captioning dataset. We use the public version processed by~\citep{lin2014microsoft}, where images with missing category information have been filtered out. This results in 122K labeled images by combining the training and validation splits. We randomly sample 5K images as the query set, with the remaining images forming the database, and then randomly sample 10K images from the database for training.

\paragraph{License.} 
ImageNet is released under a non-commercial license, and the use of the dataset is restricted to research and educational purposes. Users must apply for access and agree to the ImageNet Terms of Use.\footnote{\url{https://image-net.org/download}} CIFAR-10 is made publicly available by the University of Toronto under the MIT License. This permits free use, modification, and distribution of the dataset for both research and commercial purposes.\footnote{\url{https://www.cs.toronto.edu/~kriz/cifar.html}} For MSCOCO, the annotations are provided under the Creative Commons Attribution 4.0 License (CC BY 4.0), and the use of the images must comply with the Flickr Terms of Use.\footnote{\url{https://www.flickr.com/help/terms}} The dataset is released for academic and research use.

\subsection{Hyperparameter Tuning}

\begin{figure}[t]
    \centering
    \begin{subfigure}[t]{0.33\textwidth}
        \centering
        \includegraphics[width=\textwidth]{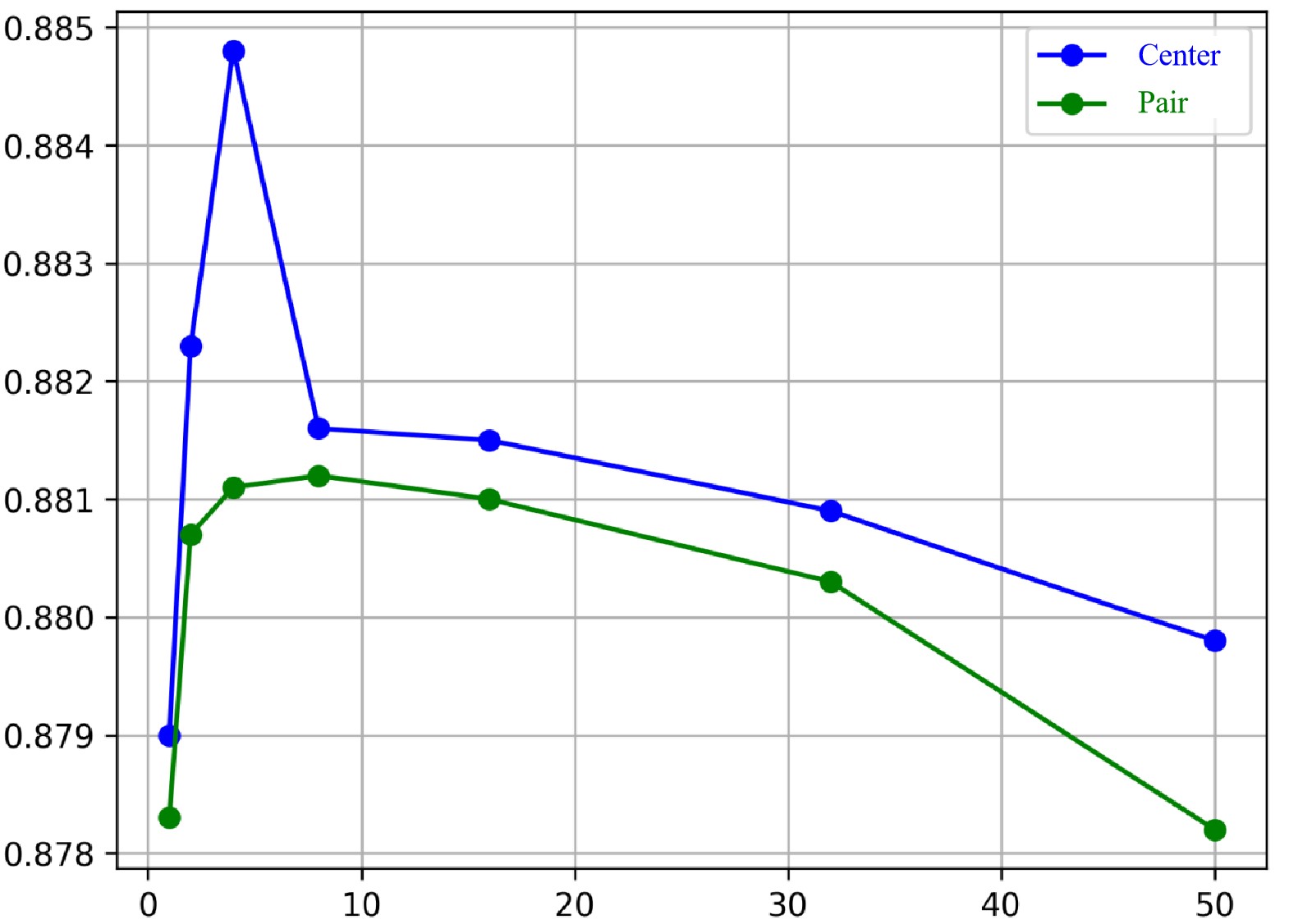}
        \caption{$\lambda_1$ Tuning Effects}
    \end{subfigure}\hfill
    \begin{subfigure}[t]{0.33\textwidth}
        \centering
        \includegraphics[width=\textwidth]{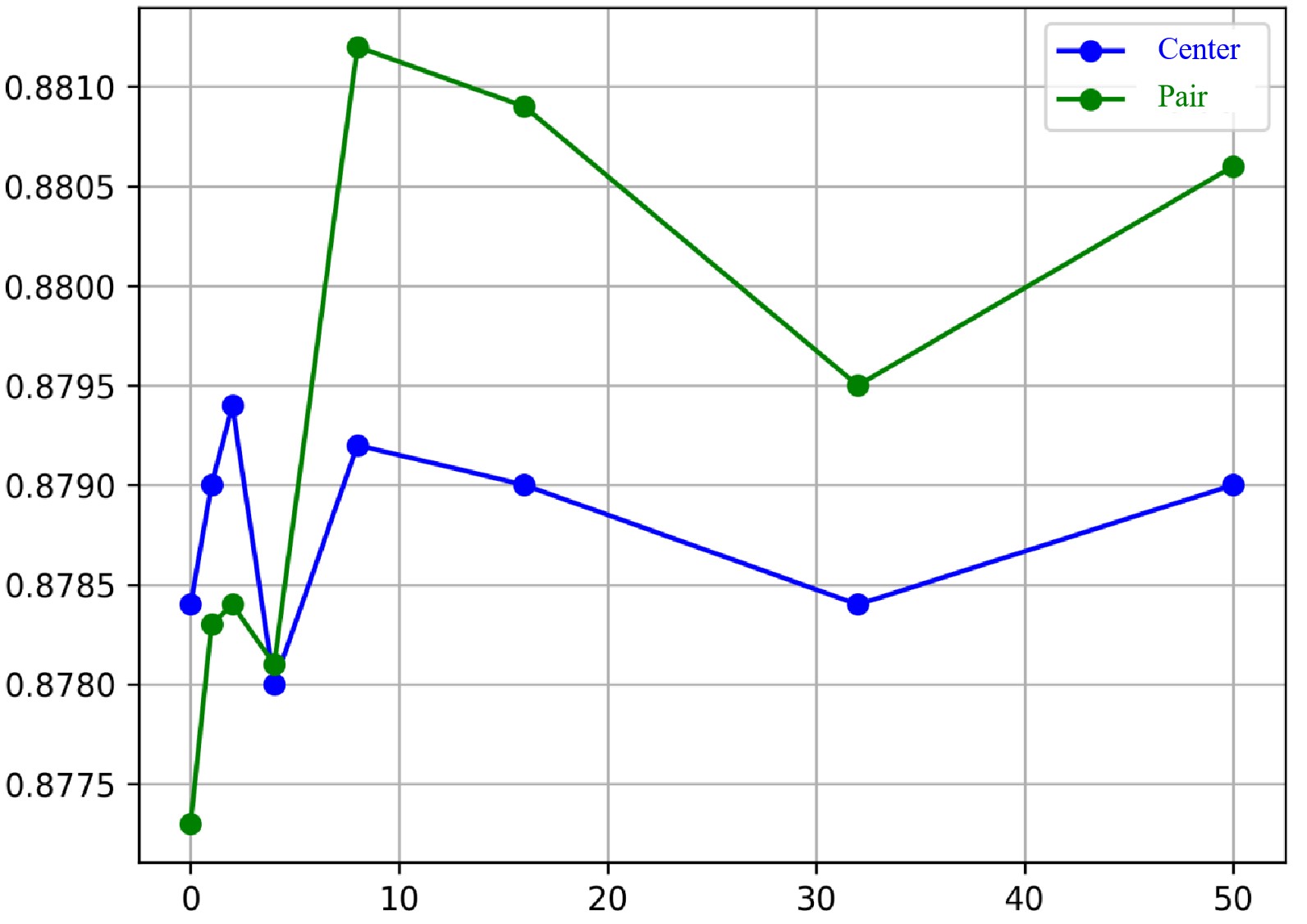}
        \caption{$\lambda_2$ Tuning Effects}
    \end{subfigure}\hfill
    \begin{subfigure}[t]{0.33\textwidth}
        \centering
        \includegraphics[width=\textwidth]{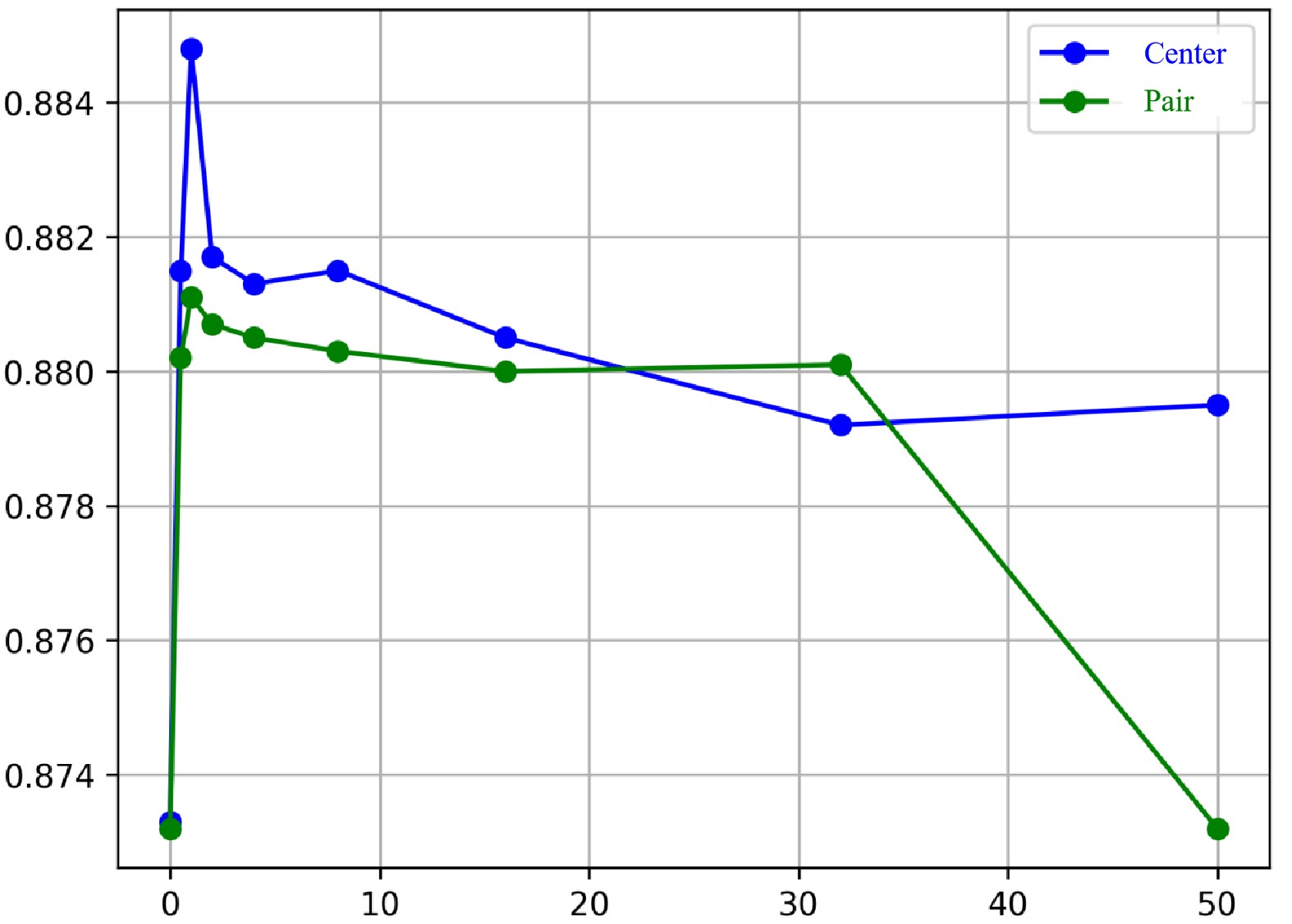}
        \caption{$\lambda_3$ Tuning Effects}
    \end{subfigure}
    \caption{Impact of Hyperparameter Tuning on Model Performance mAP for ImageNet100 with 16-bit configuration.}
    \label{fig:hyperparameters}
\end{figure}

Figure~\ref{fig:hyperparameters} illustrates a hyperparameter tuning study on the ImageNet dataset with 16-bit hash codes, varying one of \(\lambda_1\), \(\lambda_2\), or \(\lambda_3\) while fixing the others, as shown in subfigures (a), (b), and (c). In the setup, \(\lambda_1\) and \(\lambda_2\) weigh the center-based and pairwise-based losses, respectively, while \(\lambda_3\) adjusts mutual learning intensity. We report the best converged mAP for the center-based (blue) and pairwise-based (green) branches.

Key findings include: (1) Subfigures (a) and (b) reveal a dominant-auxiliary dynamic: when \(\lambda_1 \gg \lambda_2\), value1 outperforms value2, and vice versa when \(\lambda_2 \gg \lambda_1\). Configurations with the center-based branch dominating (\(\lambda_1 > \lambda_2\)) yield better overall performance than pairwise-dominated setups (\(\lambda_2 > \lambda_1\)). Thus, in our network architecture, we adopt the stronger center-based branch as the primary component, with the pairwise branch serving to fine-tune its performance. (2) Both \(\lambda_1\) and \(\lambda_2\) exhibit unimodal performance trends in pairwise and center-based  value, peaking within \([1, 10]\). (3) For \(\lambda_3\), subfigure (c) shows optimal performance at \(\lambda_3 = 1\); higher values degrade mAP, indicating excessive coupling harms learning. The best configuration is \(\lambda_1 = 4\), \(\lambda_2 = 1\), \(\lambda_3 = 1\).

\begin{figure}[t]
    \centering
    \begin{subfigure}[t]{0.33\textwidth}
        \centering
        \includegraphics[width=\textwidth]{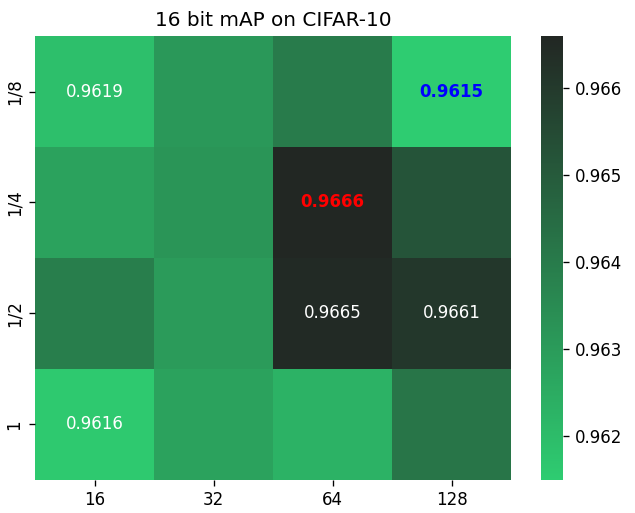}
        \caption{16bit on CIFAR-10}
    \end{subfigure}\hfill
    \begin{subfigure}[t]{0.33\textwidth}
        \centering
        \includegraphics[width=\textwidth]{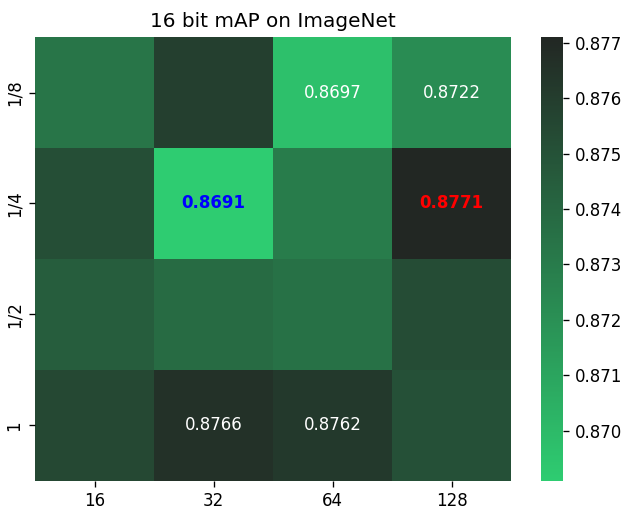}
        \caption{16bit on ImageNet}
    \end{subfigure}\hfill
    \begin{subfigure}[t]{0.33\textwidth}
        \centering
        \includegraphics[width=\textwidth]{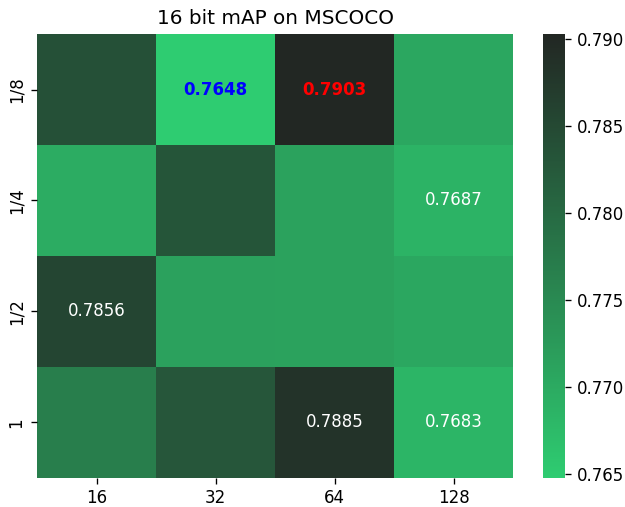}
        \caption{16bit on MSCOCO}
    \end{subfigure}
    \caption{Impact of MoH parameter tuning on model performance at 16-bit code length across ImageNet, MSCOCO, and CIFAR-10 datasets. In heatmaps, darker colors indicate higher values, while lighter colors represent lower values. The maximum and minimum values are highlighted in red and blue, respectively.}
    \label{fig:mohparameters}
\end{figure}

\section{Ablation Study and Further Analysis}
\subsection{Comparison with Traditional Mutual Learning}

Traditional Deep Mutual Learning (DML) \citep{zhang2018deep,wu2019mutual,zhao2021deep,zhao2021novel} typically employs two separate branches, where each branch contains an independently initialized and trained deep neural network (DNN). While this design allows for mutual supervision between diverse learners, it also limits the potential for fine-grained interaction between the learned representations, especially in the context of hashing where compact and consistent binary codes are desired.

In contrast, our method adopts a shared-backbone design with two branches operating on the same DNN. This encourages closer interaction and more effective information sharing between the branches, thereby facilitating the generation of more consistent and semantically aligned hash codes. The underlying idea is to enforce mutual guidance without introducing significant representational discrepancies caused by separate networks.

We evaluate both settings — one with a single shared DNN (denoted as 1DNN+MoH), and one with two independent DNNs (denoted as 2DNN+MoH) — across three benchmark datasets: ImageNet, MSCOCO, and CIFAR-10. As shown in Table~\ref{tab:dml_comparison}, our shared DNN design consistently outperforms the traditional dual-DNN setup across almost all bit lengths and datasets.

\begin{table}[t]
\small
\centering
\caption{Performance of 1DNN+MoH and 2DNN+MoH on ImageNet, MSCOCO, and CIFAR-10 across all bits. Best results are \textbf{bolded}.}
\begin{tabular}{l*{9}{c}}
\toprule
 & \multicolumn{3}{c}{\textbf{ImageNet}} & \multicolumn{3}{c}{\textbf{MSCOCO}} & \multicolumn{3}{c}{\textbf{CIFAR-10}} \\
\cmidrule(lr){2-4} \cmidrule(lr){5-7} \cmidrule(lr){8-10}
\textbf{Method} & \textbf{16-bit} & \textbf{32-bit} & \textbf{64-bit} & \textbf{16-bit} & \textbf{32-bit} & \textbf{64-bit} & \textbf{16-bit} & \textbf{32-bit} & \textbf{64-bit} \\
\midrule
2DNN+MoH & 0.8601 & 0.8859 & 0.8996 & 0.7374 & 0.8217 & 0.8623 & 0.9632 & 0.9650 & 0.9647 \\
\textbf{1DNN+MoH} & \textbf{0.8744} & \textbf{0.8975} & \textbf{0.9062} & \textbf{0.7903} & \textbf{0.8675} & \textbf{0.8727} & \textbf{0.9665} & \textbf{0.9657} & \textbf{0.9658} \\
\bottomrule
\label{tab:dml_comparison}
\end{tabular}
\end{table}

\subsection{MoH Module Analysis}

To better understand the efficacy of our proposed Mixture-of-Hash-Experts (MoH) module, we conduct ablation studies targeting three core components: the design of hashing experts, expert sharing, and the role of the softmax mechanism. Table~\ref{tab:moh_analysis} summarizes the experimental results across three datasets.

\textbf{Design of Hashing Experts vs. Traditional Experts.} Traditional Mixture of Experts (MoE) \citep{chen2023adamv,chen2022towards,riquelme2021scaling,shazeer2017outrageously} typically employs two-layer MLPs with ReLU activations as experts, designed for general-purpose representation transformation. In contrast, our MoH replaces these with specialized hashing experts, which directly map the feature dimension to the hash bit dimension — effectively acting as task-specific hashing layers.

We compare two baselines: the traditional MoE expert, which uses a two-layer MLP with hidden ReLU as commonly seen in the MoE literature, and the traditional hash expert, which consists of a single linear projection layer without non-linearity, as typically employed in hashing methods.

Our design strikes a balance: it retains the structure of two-layer MLPs but aligns their output directly to binary codes, offering greater representational power while preserving hash compatibility. As shown in the table, both traditional variants perform worse than our method, especially on MSCOCO (e.g., 0.8675 vs. 0.8472 for 32-bit).

\textbf{Expert Sharing Across Branches.} We adopt a shared expert design across branches in MoH to encourage consistent hashing and reduce redundancy. To verify its effectiveness, we compare with a variant where each branch has separate (unshared) experts.

Results show that unshared experts degrade performance on all datasets. For instance, on ImageNet at 32-bit, shared experts achieve 0.8975 vs. 0.8958 with unshared experts, confirming that expert sharing enhances generalization and code consistency.

\textbf{Impact of Removing Softmax Gate.} Unlike traditional MoE which utilizes softmax to weigh expert contributions, we remove softmax and instead allow parallel supervision from all experts. This simplifies optimization and encourages more diverse expert behaviors. As Table~\ref{tab:moh_analysis} shows, removing softmax leads to consistent improvements: for instance, on ImageNet (64-bit), performance rises from 0.9001 (with softmax) to 0.9062 (ours).

\begin{table}[t]
\small
\centering
\caption{Performance comparison of different methods on ImageNet, MSCOCO, and CIFAR-10 across all bits. Traditional MoE experts typically employ two-layer MLPs with ReLU activations. Traditional hashing expert consists of a single linear projection layer without non-linearity, as commonly used in hashing-based methods. Unshared expert indicates that the two branches do not share experts. "Trad" in this table means traditional. Best results are \textbf{bolded}.}
\begin{tabular}{l*{9}{c}}
\toprule
 & \multicolumn{3}{c}{\textbf{ImageNet}} & \multicolumn{3}{c}{\textbf{MSCOCO}} & \multicolumn{3}{c}{\textbf{CIFAR-10}} \\
\cmidrule(lr){2-4} \cmidrule(lr){5-7} \cmidrule(lr){8-10}
\textbf{Method} & \textbf{16-bit} & \textbf{32-bit} & \textbf{64-bit} & \textbf{16-bit} & \textbf{32-bit} & \textbf{64-bit} & \textbf{16-bit} & \textbf{32-bit} & \textbf{64-bit} \\
\midrule
Trad MoE Expert & 0.8762 & 0.8862 & 0.8942 & 0.7730 & 0.8472 & 0.8605 & 0.9649 & 0.9653 & 0.9634 \\
Trad Hash Expert & 0.8663 & 0.8872 & 0.8996 & 0.7584 & 0.8533 & 0.8658 & 0.9638 & 0.9641 & 0.9649 \\
Unshared Expert & 0.8728 & 0.8958 & 0.9031 & 0.7882 & 0.8657 & 0.8701 & 0.9658 & 0.9659 & 0.9657 \\
With Softmax & 0.8679 & 0.8907 & 0.9001 & 0.7793 & 0.8526 & 0.8686 & 0.9625 & 0.9639 & 0.9647 \\
\textbf{MoH} & \textbf{0.8744} & \textbf{0.8975} & \textbf{0.9062} & \textbf{0.7903} & \textbf{0.8675} & \textbf{0.8727} & \textbf{0.9665} & \textbf{0.9657} & \textbf{0.9658} \\
\bottomrule
\end{tabular}
\label{tab:moh_analysis}
\end{table}

\subsection{MoH Parameter Tuning}

We investigate the influence of two hyperparameters in the MoH module: the number of total experts (horizontal axis) and the activation ratio, i.e., the proportion of experts selected per input (vertical axis). As shown in Figure~\ref{fig:mohparameters}, each subfigure presents the model's 16-bit mAP on CIFAR-10, ImageNet, and MSCOCO, respectively.

Overall, MoH demonstrates stable performance across a wide range of settings, but appropriate tuning of these parameters can yield noticeable improvements. On ImageNet, the best performance (0.8771 mAP) is achieved when using 64 experts with a 1/4 activation ratio, indicating a balanced trade-off between diversity and sparsity. On CIFAR-10, performance remains consistently high, with the best result (0.9666 mAP) also occurring at 64 experts and a 1/4 ratio. For MSCOCO, the highest mAP (0.7903) is observed when activating 1/8 of 64 experts, suggesting that a smaller number of activated experts may be more effective for denser datasets.

It is also notable that overly low expert counts (e.g., 16) or excessively sparse activation (e.g., 1/8 on CIFAR-10) tend to hurt performance, likely due to insufficient model capacity or representational bottlenecks. These results suggest that MoH benefits from a moderate number of diverse experts, with partial activation to maintain efficiency and specialization.

\section{Limitations}

While our proposed Mutual Learning for Hashing (MLH) consistently demonstrates strong performance across diverse datasets and settings, several complementary aspects remain open for further exploration. First, the dual-branch architecture and the MoH module introduce a more intricate interaction mechanism in the hash code generation process, which makes the interpretability of individual bits less straightforward and difficult to analyze directly. Second, like many deep hashing methods, MLH may encounter practical limitations in resource-constrained environments due to its reliance on deep neural networks. However, as numerous methods exist to improve the deployability of deep neural networks, this limitation has been less significant. These considerations highlight valuable directions for future work while underscoring the effectiveness of our method.

%\bibliographystyle{unsrtnat}
%\bibliography{references.bib}

 \end{document}